%% file: my.tex
\documentclass[journal,twoside]{IEEEtran}

\usepackage{graphicx}
\usepackage{url}
\usepackage{soul}
\usepackage{amsmath}
\usepackage{amssymb}

\usepackage{color}
\sethlcolor{yellow}
\usepackage{cite}
\soulregister\cite7

\ifCLASSINFOpdf
\else
\fi
\begin{document}
%
\title{Structure-Aware Network for Lane Marker Extraction with Dynamic Vision Sensor}
%
%
%


\author{Wensheng~Cheng*,
	    Hao~Luo*,
        Wen~Yang,~\IEEEmembership{Senior Member, IEEE,}
        Lei~Yu,~\IEEEmembership{Member, IEEE,}
        and~Wei~Li

\thanks{W. Cheng, H. Luo, W. Yang, L. Yu are with the School of Electrical Information, Wuhan University, Wuhan 430070, China(e-mail: cwsinwhu@whu.edu.cn; luohaowhu@whu.edu.cn; yangwen@whu.edu.cn; ly.wd@whu.edu.cn).}
\thanks{W. Li is with Shanghai Baolong Automotive Corporation, China(e-mail: liwei@chinabaolong.net).}
\thanks{*Equal contribution.}
}



\maketitle

\begin{abstract}

Lane marker extraction is a basic yet necessary task for autonomous driving. Although past years have witnessed major advances in lane marker extraction with deep learning models, they all aim at ordinary RGB images generated by frame-based cameras, which limits their performance in extreme cases, like huge illumination change. To tackle this problem, we introduce Dynamic Vision Sensor (DVS), a type of event-based sensor to lane marker extraction task and build a high-resolution DVS dataset for lane marker extraction. We collect the raw event data and generate 5,424 DVS images with a resolution of 1280$\times$800 pixels, the highest one among all DVS datasets available now. All images are annotated with multi-class semantic segmentation format. We then propose a structure-aware network for lane marker extraction in DVS images. It can capture directional information comprehensively with multidirectional slice convolution. We evaluate our proposed network with other state-of-the-art lane marker extraction models on this dataset. Experimental results demonstrate that our method outperforms other competitors. The dataset is made publicly available, including the raw event data, accumulated images and labels. 

\end{abstract}


%
\IEEEpeerreviewmaketitle

\input{source/01intro.tex}

\input{source/02relate.tex}
\input{source/03data.tex}

\input{source/04method.tex}

\input{source/05exp.tex}

\input{source/06conc.tex}

\ifCLASSOPTIONcaptionsoff
  \newpage
\fi



\bibliographystyle{IEEEtran}
\bibliography{IEEEabrv,dvs,segmentation}





\end{document}

%% file: source/01intro.tex
\section{Introduction}
\IEEEPARstart{A}{utonomous} driving has received much attention in both academia and industry. It aims to perceive the surrounding environment via various sensors and make the decision correspondingly. It involves many tasks, like pedestrian detection \cite{zhu2016traffic,li2018scale,tian2015pedestrian}, traffic mark recognition and lane marker extraction. Lane marker extraction enables the car to follow the lanes precisely. Many applications are based on the task, including trajectory planning and lane departure. It becomes a key aspect for autonomous driving.

\begin{figure}[tbp]
	\begin{minipage}[b]{1.0\linewidth}
		\centering
		\centerline{\includegraphics[width=8cm]{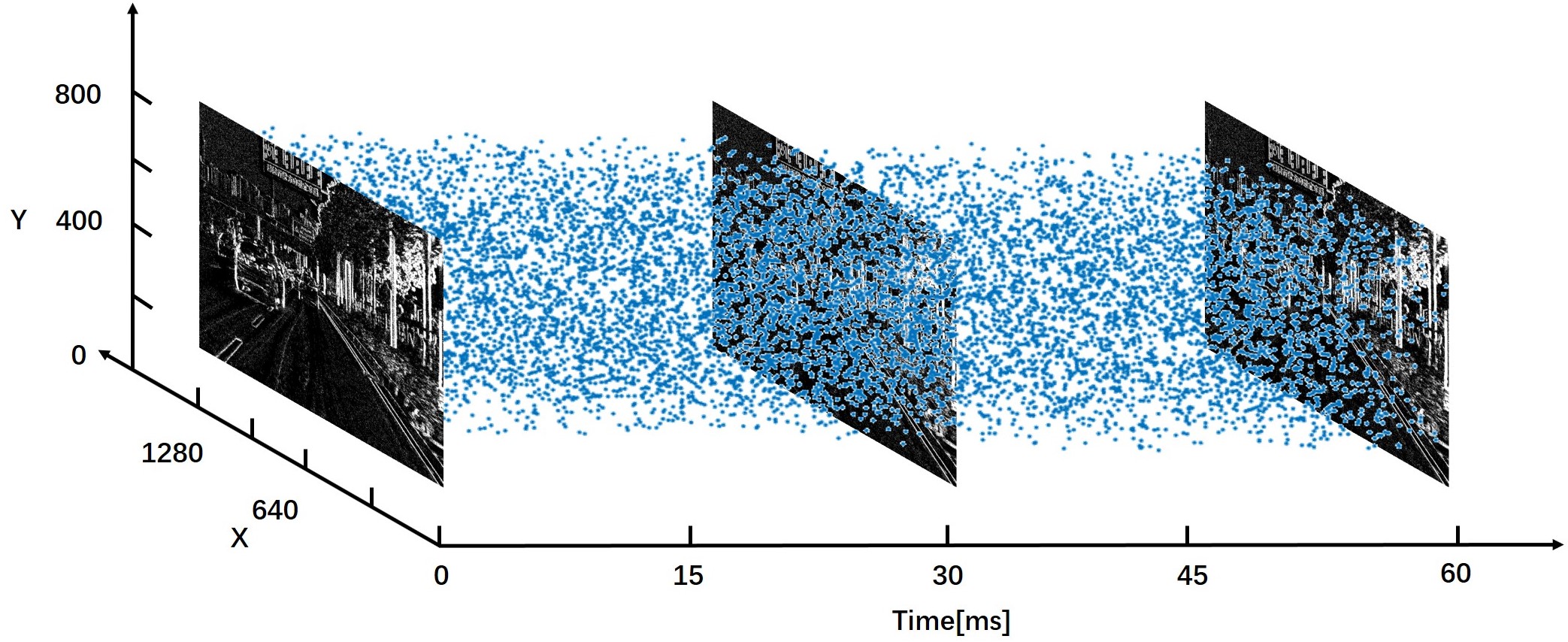}}
	\end{minipage}
	\caption{Visualization of the event output in	space-time. Blue dots represent individual asynchronous events.}
	\label{fig:theory}
	\vspace{-0.5cm}
\end{figure}

For lane marker extraction task, many methods have been proposed by researchers. These methods include handcrafted features and heuristic algorithms~\cite{borkar2012novel, deusch2012random, hur2013multi, jung2013efficient,tan2014novel, wu2014lane}, and end-to-end Convolutional Neural Network (CNN) models~\cite{gopalan2012learning,kim2014robust,huval2015empirical,he2016accurate,li2017deep,lee2017vpgnet}. Although promising results have been achieved by them, there are still challenging scenes in practice.

In fact, various extreme and complex scenes could happen. Take an example, fast changing light or low illumination condition would severely influence the performance of these methods. Under these conditions, general frame-based cameras are not able to capture these scenes clearly, so these methods cannot work well with the bad input \cite{binas2017ddd17}.
Therefore, we resort to DVS. DVS camera only produces data when photometric changes occur at certain pixels in the sensor, and each pixel operates in an asynchronous and independent way. DVS has shown its potential for visual tasks in recent years \cite{valeiras2018event,cohen2018spatial,camunas2017event}. The event output visualization is shown in Fig.\ref{fig:theory}. There are two key characteristics: low latency and high dynamic range. Latency is based on the sensor sampling rate and data processing time. Since DVS transmits data with events, which denote illumination change, it has a latency of microseconds ($\mu s$), compared with 50-200 milliseconds ($ms$) of standard cameras \cite{mueggler2017event}. With such low latency, DVS can sense the environment and capture images much faster than standard cameras. This property can alleviate the influence efficiently caused by motion blur, which is a troublesome problem for frame-based cameras. Besides, with much shorter response time brought by low latency, it also makes the autonomous cars much more agile than others.
\begin{figure*}[t]
	\begin{minipage}[b]{1.0\linewidth}
		\centering
		\centerline{\includegraphics[width=16cm]{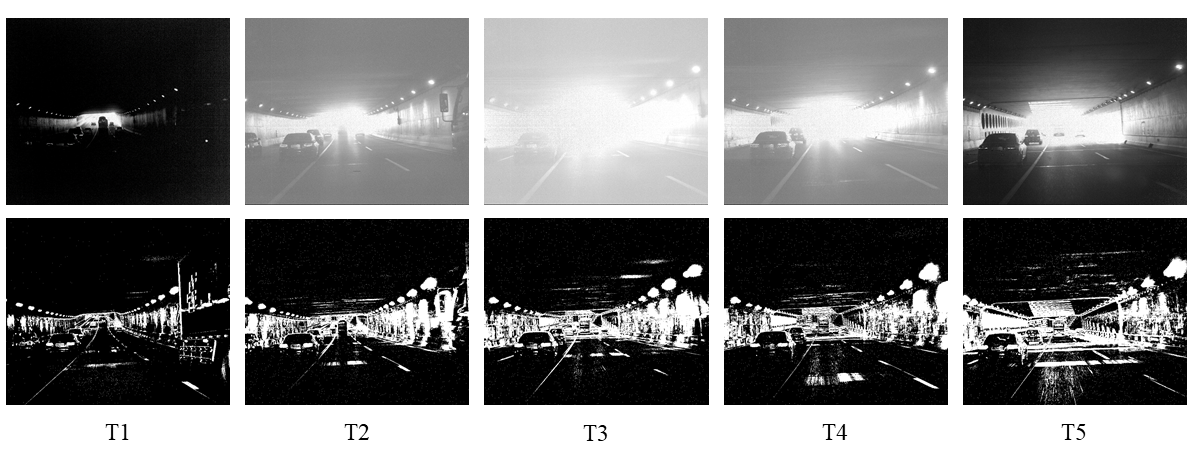}}
	\end{minipage}
	\caption{The process of car's coming out of the tunnel (chronological order: $T1<T2<T3<T4<T5$). The first row shows gray images captured by frame-based camera. The second row shows corresponding DVS images captured the same moment. Traditional cameras are largely affected by the sudden light change due to the low dynamic range, while DVS doesn't suffer from that with much higher dynamic range. }
	\label{fig:advantage}
\end{figure*}

About dynamic range, the typical dynamic range of DVS is 130 dB v.s. 60 dB for general frame-based ones, which is 7 orders of magnitude larger \cite{mueggler2017event}. This characteristic enables it to tackle extreme scenes, like large illumination changes. Suppose a car is going through the tunnel, the moments it enters and leaves the tunnel would result in such dramatic illumination change and corresponding images would become highly dark or light. This makes general frame-based cameras almost impossible to recognize lanes from these images. But for DVS, lanes are still clear due to the high dynamic range, as shown in Fig.\ref{fig:advantage}.

Besides, with the event stream data, semi-dense images are generated by DVS. So only pixels whose brightness changes as a result of relative movement would be shown on the DVS output image. These pixels usually come from pedestrians, traffic marks, cars and pedestrians. They are the key objects for autonomous driving. Road surface, sky and other background information would be removed.

A possible problem for utilizing DVS for lane marker extraction task is the DVS image resolution. The images generated by ordinary DVS only have a resolution about 240$\times$180 pixels, which contains a few details.

For reasons above, we construct a high-resolution DVS dataset for lanE marker exTraction (DET)\footnote{Dataset website is \url{https://spritea.github.io/DET/}} with a high-resolution DVS camera, CeleX V. There are 5,424 DVS images of 1280$\times$800 pixels with corresponding labels. Note that we also provide the raw event data for those algorithms taking use of event data directly \cite{sironi2018hats,lagorce2017hots}. The partition is: 2,716 images as training set, 873 images as validation set and 1,835 images as test set. Per-pixel labels with distinguishable lanes are provided. This is because many advanced models for lane marker extraction are based on semantic segmentation technique, which requires this kind of label. As far as we know, this is the first dataset for lane marker extraction with DVS images. It's also the first DVS dataset with such high resolution.

For lane marker extraction task, contextual structural information is of significant importance. The information is critical for building contextual relation to recognize objects with long continuous shape. It matters even more for lane marker extraction with DVS images, due to the lack of appearance information. Existing state-of-the-art methods for lane marker extraction task are generally based on CNN, which suffers from the lack of suitable capacity for capturing contextual information owing to the limited empirical receptive field \cite{zhou2014object}. 

To handle the problem, some researchers try to use global pooling layer to introduce global context \cite{szegedy2015going,liu2015parsenet}. However, this is insufficient for complex environment, because it may lose the spatial relation and cause ambiguity. Others adopt Recurrent Neural Network (RNN) to pass information along each row or column and capture contextual structural information \cite{visin2015renet,bell2016inside}. But in one RNN layer, each pixel position could only receive information from the same row or column. \cite{pan2018spatial} proposes Spatial CNN (SCNN), which generalizes traditional deep layer-by-layer convolutions to slice-by-slice convolutions within feature maps, thus enabling message passing between pixels across rows and columns in a layer. The strategy is efficient and achieves state-of-the-art performance on public dataset. But it only extracts structural information along the horizontal and vertical directions, without considering other directions. We argue that the diagonal direction also matters, since most lanes present no exact horizontal or vertical shape in the front-view DVS images. 

We then propose the structure-aware network (SANet) for lane marker extraction in DVS images. It's based on the Multidirectional Slice Convolution (MSC) module introduced in this paper, which can capture structural information along the horizontal, vertical and diagonal directions. This helps the network extract lanes more accurately. We then compare our proposed network with other state-of-the-art models on DET and report the results. Experimental results demonstrate that our method outperforms other competitors.

Compared with the conference version \cite{det} which focuses on the dataset only, this paper has further studied the method for lane marker extraction and proposed the SANet for the task. In summary, our contributions are:
\begin{itemize}
	\item We provide a DVS dataset for lane marker extraction, including the raw event data and accumulated images with labels. To our knowledge, DET is the first DVS dataset for this task and the first DVS dataset with such high resolution images of 1280$\times$800 pixels.
	\item We propose the SANet for lane marker extraction task in DVS images. It is based on the MSC module which can capture the structural information of DVS images more comprehensively.
	\item We evaluate our network with other state-of-the-art models on DET and report the results. Experimental results show that our method outperforms other competitors.
\end{itemize}


%% file: source/02relate.tex
\section{Related work}

\subsection{Event Camera Dataset} 
\textbf{Synthesized Dataset.} A Dynamic and Active-pixel Vision sensor (DAVIS) dataset and corresponding simulator have beed proposed by \cite{mueggler2017event}. DAVIS consists of an event-based sensor and a global-shutter camera. The dataset is collected with event-based sensor in various synthetic and real environments. It consists of global-shutter intensity images with asynchronous events, and movement with pose parameters. It contains lots of scenes, like office, outdoors, urban and wall poster. The purpose of the dataset is for visual odometry, pose estimation, and SLAM. The resolution of the dataset image is 240$\times$180 pixels.

\textbf{Classification Dataset.} CIFAR10-DVS \cite{li2017cifar10} is an event-stream dataset for object classification. 10,000 frame-based images that come from CIFAR-10 dataset are converted into 10,000 event streams with an event-based sensor, whose resolution is 128$\times$128 pixels. The dataset has an intermediate difficulty with 10 different classes. The repeated closed-loop smooth (RCLS) movement of frame-based images is adopted to implement the conversion. Due to the transformation, they produce rich local intensity changes in continuous time which are quantized by each pixel of the event-based camera.

\textbf{Recognition Dataset.} A series of DVS benchmark datasets are released in \cite{hu2016dvs}. Visual video benchmarks, including object recognition, action recognition and object tracking are converted into spiking neuromorphic datasets. A DAViS240C camera of 240$\times$180 pixels resolution is adopted to record in the process. Four classic dynamic datasets are transformed: Tracking Dataset \cite{tracking}, the VOT challenge 2015 Dataset \cite{vot2015}, the UCF-50 Action Recognition Dataset \cite{reddy2013recognizing} and the Caltech-256 Object Category Dataset \cite{griffin2007caltech}.

\textbf{Driving Dataset.} DDD17 \cite{binas2017ddd17} is an open dataset of annotated DAVIS driving recordings. It contains 12 hour record of a DAVIS sensor with a resolution of 346$\times$260 pixels.

It is a city and highway driving in all kinds of weather conditions, along with GPS position and vehicle speed. The data also contains driver steering, brake, and throttle captured from the car's on-board diagnostics interface. Since there are data from various devices and sensors, it is very helpful for autonomous driving task. 

DVS datasets listed above are proposed for general computer vision or robotic control tasks. None of them targets the lane marker extraction task. Besides, event-based images in these datasets only have a low spatial resolution, like 240$\times$180 pixels or 128$\times$128 pixels. The low resolution secerely limits algorithms' performance on these datasets.
\subsection{Lane Dataset}
\textbf{Caltech Lanes Dataset.} This dataset \cite{aly2008real} is released in 2008 as an early one. It contains clips under various situations, including straight and curved streets, different types of urban streets, with/without shadows. All visible lanes are labeled in four clips. There are totaling 1,224 labeled frames with 4,172 marked lanes. 

\textbf{TuSimple Dataset.} This dataset \cite{tusimple} contains 6,408 labeled images. They are partitioned into training set of 3,626 images and test set of 2,782 images. These images are obtained under medium and good weather condition. There are highway roads with a different number of lanes, like 2 lanes, 4 lanes or more. For each image, 19 previous frames are also provided without annotation.

\textbf{CULane Dataset.} This dataset \cite{pan2018spatial} consists of 133,235 frames extracted from 55 hours of video. The dataset is split into training set of 88,880 images, validation set of 9,675 images, and test set of 34,680 images. The resolution of these undistorted images is 1640$\times$590 pixels. The test set contains normal and other challenging categories, like shadow and crowded scenes.

These lane datasets are all based on RGB images generated by frame-based cameras. Illumination changes and motion blur would affect model's performance based on theses images seriously, which should definitely be avoided in real traffic situation.
\begin{figure*}[t]
	\begin{minipage}[b]{1.0\linewidth}
		\centering
		\centerline{\includegraphics[width=16cm]{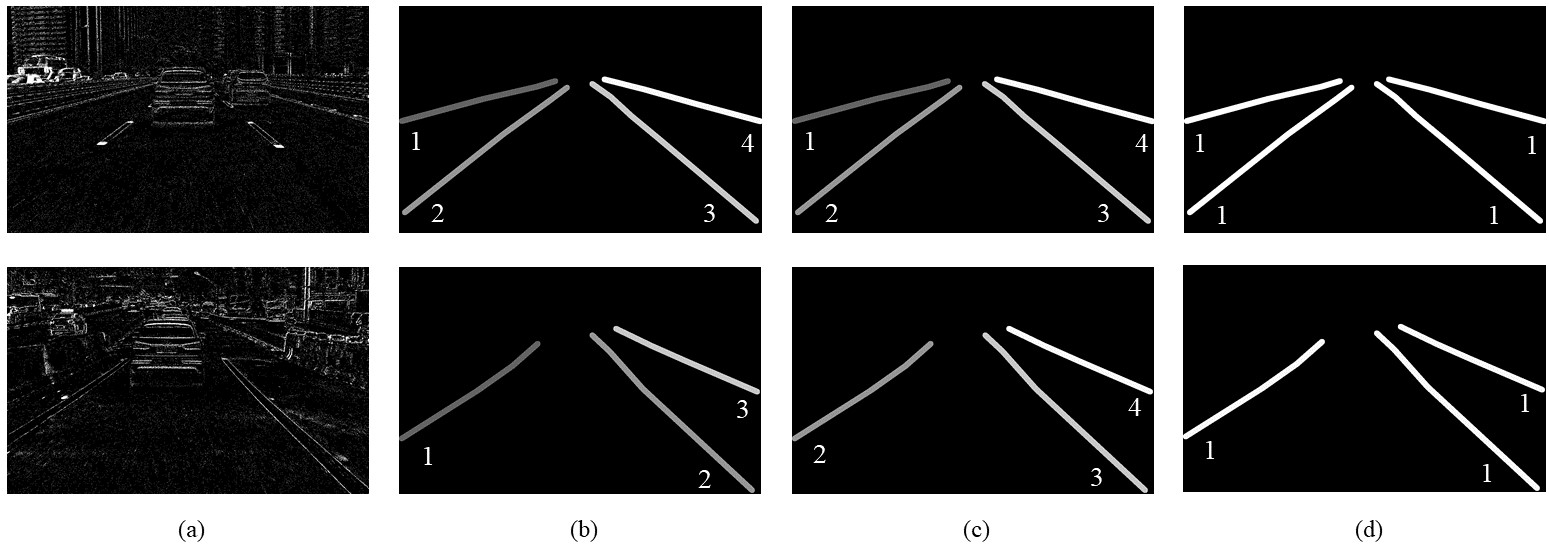}}
	\end{minipage}
	\caption{Comparison of different label formats. (a) shows input images. (b) shows the label format that sets a fixed order and annotates lanes from left to right. (c) shows our label format based on the relative distance between lane and event camera. (d) shows the binary label format. For the left lane most close to event camera, it looks similar in different images and should be annotated with same label. (b) gives it a label of 2 in the image above, but gives it a label of 1 in the image below. Our format (c) annotates it exactly in both images.}
	\label{fig:label}
\end{figure*}
\subsection{Lane Marker Extraction Methods}
\label{sec:lane}
Lane marker extraction task has received continuous attention recently. Hence, researchers have proposed various methods to solve this problem. These methods could be divided into two categories generally, traditional methods and deep learning methods.

\textbf{Traditional Methods.} Traditional methods usually consist of handcrafted features and heuristic algorithms. \cite{chiu2005lane} uses statistical method to find out a color threshold. Then it utilizes color-based segmentation method to find out the lane boundary. \cite{loose2009kalman} combines Kalman filter and particle filter together to process 3D information, which is obtained from stereo vision or image radar. \cite{teng2010real} integrates multiple cues, including bar filter, color cue, and Hough Transform. It adopts particle filtering technique to realize lane tracking. \cite{lopez2010robust} employs ridge features to this problem. It also adapts RANSAC \cite{fischler1981random} to fit a parametric model of a pair of lane lines to the image features. \cite{zhou2010novel} presents a robust lane detection algorithm based on geometrical model and Gabor filter. \cite{aly2008real} adopts selective oriented Gaussian filters and RANSAC to fit Bezier Splines.

\textbf{Deep Learning Methods.} Deep learning technique, especially CNN, has shown its superiority on image processing field recently. Many lane marker extraction methods based on CNN have been proposed. \cite{gopalan2012learning} uses a pixel-hierarchy feature descriptor to model the contextual information shared by lane markings with the surrounding road region. It also adopts a robust boosting algorithm to select relevant contextual features for detecting lane markings. \cite{kim2014robust} combines CNN with RANSAC algorithm to detect lanes. It only takes use of CNN model when the traffic scene is complicated and RANSAC isn't able to deal with it. \cite{huval2015empirical} takes existing CNN models to perform lane and vehicle detection while running at frame rates required for a real-time system. \cite{he2016accurate} proposes the DVCNN strategy, which utilizes both front-view and top-view image to exclude false detections and non-club-shaped structures respectively. \cite{li2017deep} develops a multitask deep convolutional network, which detects the presence and the geometric attributes of the target at the same time. It also adopts a recurrent neuron layer to detect lanes. LaneNet \cite{lane_net} casts the lane marker extraction problem as an instance segmentation problem. A learned perspective transformation based on the image is applied without a fixed ``bird's eye view'' transformation. clustering is used to generate each lane instance. Hence it can handle scenes where lane category varies, although it cannot assign similar lanes same label.
SCNN \cite{pan2018spatial} is based on semantic segmentation backbone. It generalizes traditional deep layer-by-layer convolutions to slice-by-slice convolutions within feature map, which enables message passing between pixels across rows and columns in a layer.

Although these CNN-based methods get obvious improvement for this task, most of them adopt conventional convolutional layer directly, without exploiting structural information explicitly. This limits their performance in essence.

%% file: source/03data.tex
\section{Construction of DET}
\begin{figure*}[t]
	\begin{minipage}[b]{1.0\linewidth}
		\centering
		\centerline{\includegraphics[width=16cm]{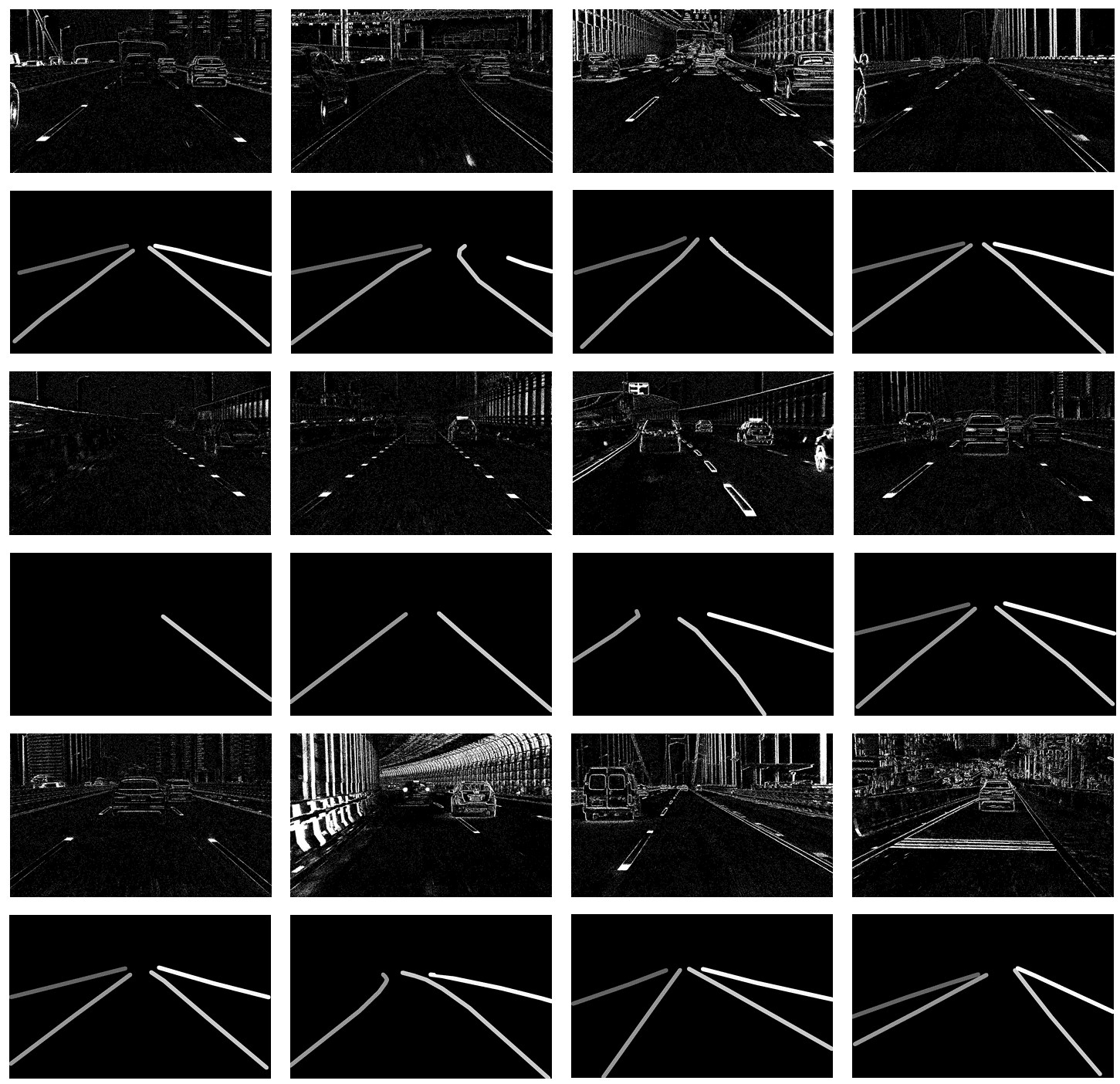}}
	\end{minipage}
	\caption{Samples of DVS images and labels in DET. First rows show various line types, including single dotted line, single solid line, parallel dotted line, parallel solid line and dotted line. Middle rows show various lane number, from 1 to 4. Last rows show various traffic scenes, urban, tunnel, bridge and overpass.}
	\label{fig:dataset}
\end{figure*}

\subsection{Data Collection}
A CeleX V DVS camera with a resolution of 1280$\times$800 pixels is adopted to collect the event data. The camera is mounted on a car in various locations driving in different time in Wuhan City. Wuhan City is a metropolis of China, so there are various and complex traffic scenes which become a challenge for lane marker extraction. 

We record over 5 hours of event stream with a sampling rate of MHz, which equals a sampling interval of $\mu s$. The raw event stream is compressed with $\triangle t=30\ ms$ along the time dimension. Fig.\ref{fig:theory} demonstrates the process. Then over 150,000 images are generated from raw event stream. 5,424 images containing various scenes are selected to annotate.


\subsection{Data Annotation}
\label{sec:da}
\textbf{Task Definition.} There are two kinds of definitions for lane marker extraction task. One is to extract lanes without discriminating between lanes, and the other one is to differentiate lanes from each other. We argue the latter is more practical, because ego lanes are labeled as different categories from other lanes, the location of the car could be decided directly, i.e., the car is between the two ego lanes. This is necessary for following application, like cruise control. But it is not easy for the former way to decide the car's specific location. Therefore, we define lane marker extraction here as extracting lanes from traffic scenes while discriminating between lanes. The definition is the same with the existing best method, SCNN \cite{pan2018spatial}.

Following the definition above, lane marker extraction is in fact a multi-class semantic segmentation problem. It classifies each pixel into $(n+1)$ categories, where $n$ denotes lane types and $1$ denotes background. Lanes with same label are supposed to be similar in some sense.

\textbf{Annotation Details.} Semantic segmentation method requires multi-class label. As the first lane marker extraction dataset with DVS camera, we choose the most representative situation of 4 lanes in the dataset, so there are 4 lanes at most in one image. This choice is the same with existing lane datasets, like CULane Dataset \cite{pan2018spatial}. Hence it's a five-class classification task. Each pixel is given one of five-class labels, i.e., $\{0, 1, 2, 3, 4\}$. $0$ is for background and others for lanes. Here comes the question, that how we decide the label for each lane.

Generally, two types of rules are used to give the specific label. The first one is to set a fixed order and give each lane a label by this order. The second one is to give lanes who have similar characteristics same label. We think the latter is better. Because the first label format is only related to the number of lanes in the image, and it does not consider lane's property at all. In this way, each lane would be labeled by a fixed order, like 1 to 4 for lanes from left to right. Then lanes with same label from different images may differ much. Fig.~\ref{fig:label} (b) shows an example. This is because the distance between DVS and those lanes with same label from different images would vary a lot during the driving process. Therefore if we adopt the format, it would have bad influence on the training process of multi-class semantic segmentation model.  

\begin{table}[tb]
	\caption{Distribution of images containing various number of lanes. One represents the image containing only one lane. }
	\begin{center}
		\begin{tabular}{l|cccc|c}
			\hline
			Statistics &One & Two&Three&Four&Total\\
			\hline
			\hline
			Quantity & 161&1,114&1,918&2,231&5,424\\
			Percentage \% & 2.97&20.54&35.36&41.13&100\\
			\hline
		\end{tabular}
		\label{tab:lane-number}
	\end{center}
	\vspace{-0.5cm}
\end{table}

For reasons above, the latter format is adopted in this paper to label images. The key aspect is the definition of similarity. Since the sizes and shapes of lanes in image mainly depend on the distance between DVS and lanes, lanes whose distances from DVS are analogous would seem similar. Therefore, for ego lanes \cite{kim2017end}, i.e., the two lanes most close to DVS, the label of left lane is set as 2, and the label of right lane is set as 3, which has nothing to do with the number of lanes in the image. Other lanes' labels are decided by their distance to the ego lanes. Fig.~\ref{fig:label} (c) shows an example. By this format, lanes with similar appearances would have same label. This is more reasonable when we regard the task as a multi-class semantic segmentation problem.

About lane width, because lanes appear in various sizes due to their real sizes and effects of lens, and even different parts of the same line present different sizes or shapes, we adopt the general label format for this task. That is representing each lane with a series of key points, which are usually on the center line of the lane. CULane \cite{pan2018spatial} and TuSimple \cite{tusimple}, two widely used lane datasets, both adopt this format. So we also follow this way. We provide data files recording the coordinates of lane key points in the dataset, and the default setting of lane width is 20 pixels. Users can set a fixed lane width for all lanes, or set different lane widths for each lane according to their needs.

\begin{figure*}[t]
	\begin{minipage}[b]{1.0\linewidth}
		\centering
		\centerline{\includegraphics[width=16cm]{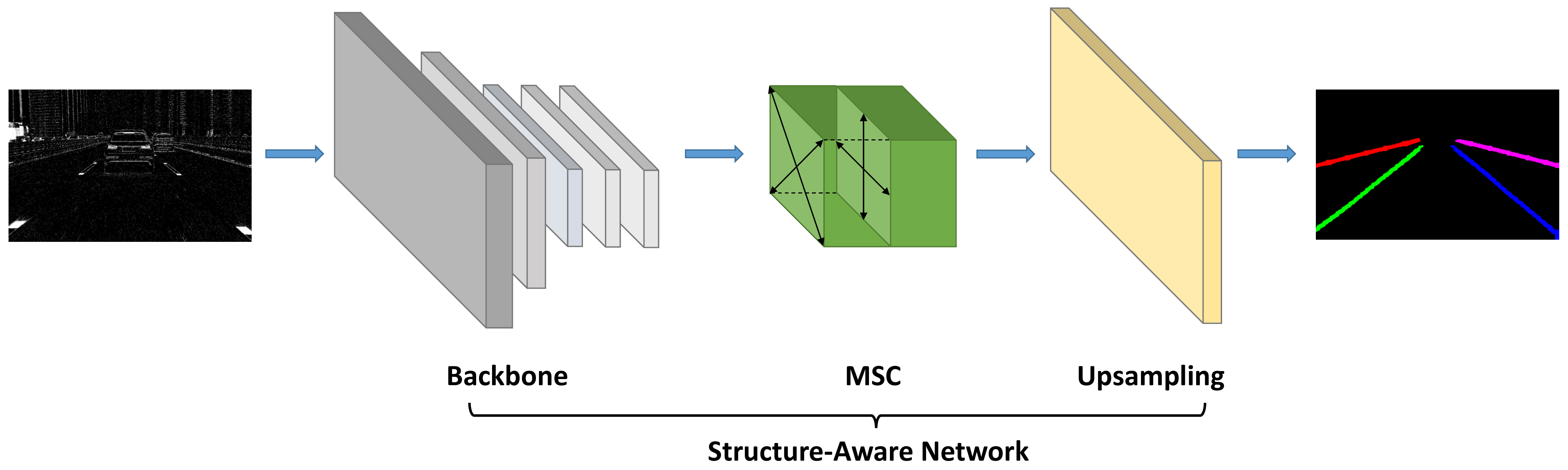}}
	\end{minipage}
	\caption{The pipeline of proposed SANet. Backbone denotes the basic network for feature extraction. MSC represents the multidirectional slice convolution module. Upsampling is used to restore the resolution of the output to be the same with the input image.}
	\label{fig:all}
\end{figure*}

For the occluded lanes, like part of one lane is occluded by a car, we would still label the occluded part and get a whole consecutive lane, not discrete lane segments. Because we adopt the key-point label format for each lane, the final label is implemented by connecting these points into a single thin lane and expanding the lane to the appointed width. So for key points on two separate segments of the same broken lane, they would be connected in the same way with those key points on one consecutive lane, which would generate a consecutive lane. Hence both broken and consecutive lane would be labeled as a whole consecutive lane. This is the same choice with the widely used CULane dataset \cite{pan2018spatial}.

For methods defining this task as extracting lanes without discriminating between them, or extracting lanes first then differentiating them in following stages, binary labels are also provided for researchers interested in this. Fig.~\ref{fig:label} (d) shows these labels. To ensure the annotation quality, we have designated tens of researchers who are experienced in related area to label these images carefully. After the initial labeling, the results were cross-checked to further improve the annotation quality.

\subsection{DET Properties}

\textbf{Data Partition.} About data partition, 1/2 of original images are randomly extracted as training set, 1/6 as validation set and 1/3 as test set.
This helps to make the distributions of these sets match approximately.
All images, including raw DVS images and filtered ones with corresponding labels, containing binary and multi-class labels would be made publicly available.

\textbf{High-resolution Image.}
The typical resolution for current DVS datasets is 346$\times$260 pixels, which is really low compared with RGB images output by frame-based cameras. If a car meets complex scenes, DVS images with such low resolution containing a little information cannot deal with the situation well. Hence, the CeleX-V DVS which was released in 2018 is adopted for our dataset. It has the highest resolution of 1280$\times$800 pixels among all DVS available now and a latency as low as 5$ns$. All images in DET have the same resolution of 1280$\times$800.

\textbf{Various Lane Types.}
The lane diversity is an important aspect for lane marker extraction dataset. Diverse lane types make the dataset more close to real world. Therefore, the dataset images contain various lane types, like single solid line, single dashed line, parallel solid line and dashed line, parallel dashed lines, etc. Note that parallel lines are labeled as one whole line for consistency. Fig.~\ref{fig:dataset} demonstrates lane type samples.

\textbf{Various Lane Number.}
For images containing different numbers of lanes, lanes would have various appearances with different distances from DVS, as explained in Sec.~\ref{sec:da}. To make the dataset more comprehensive, images including different numbers of lanes are collected by driving on roads with various numbers of carriageways. Fig.~\ref{fig:dataset} shows the samples that have different numbers of lanes. The distribution of lane number is presented in Tab.~\ref{tab:lane-number}.

\textbf{Various Traffic Scenes.}
Besides the lane diversity, the scene diversity is important as well. The reason is a robust lane marker extraction method should be able to recognize lanes under various traffic scenes. This is critical for reliable autonomous driving. Therefore, we also collect images containing various traffic scenes by driving on overpasses, bridges, tunnels, urban areas, etc. Fig.~\ref{fig:dataset} shows these traffic scenes.

\textbf{Various Camera Views.}
To simulate the real situation, we further mount DVS with different locations on our car. Under this condition, even lanes with same label in different images would differ to some extent. In fact, the intraclass variance of the dataset is increased. Although it would become more challenging for lane marker extraction models, models trained with this dataset could handle complex scenes better than those with single camera view. We think this is necessary for methods to be adopted in real traffic scenes, which might be even more complicated than the dataset.

%% file: source/04method.tex
\section{Methods}
In this section, we introduce naive slice convolution at first. Then we present the proposed Multidirectional Slice Convolution module, and explain its operation process in detail. Finally we introduce the structure-aware network based on MSC module.

\subsection{Naive Slice Convolution}
Most CNN-based lane marker extraction methods \cite{gopalan2012learning,kim2014robust,huval2015empirical,he2016accurate} make use of normal convolution layer to extract features. Although CNN has demonstrated its ability to extract semantics from raw pixels with normal convolution layer, its capability to capture spatial relation along specific directions has not been explored fully. For lane marker extraction task with DVS images, the relation is of significant importance to extract structural information, because lanes in DVS images present weak appearance coherence but strong shape prior. Besides, CNN cannot connect separated parts together automatically, which usually makes it fail when occlusion occurs. In this task, however, occlusion situations often occur when cars or pedestrains cross the lane.

\begin{figure}[t]
	\begin{minipage}[b]{1.0\linewidth}
		\centering
		\centerline{\includegraphics[width=8cm]{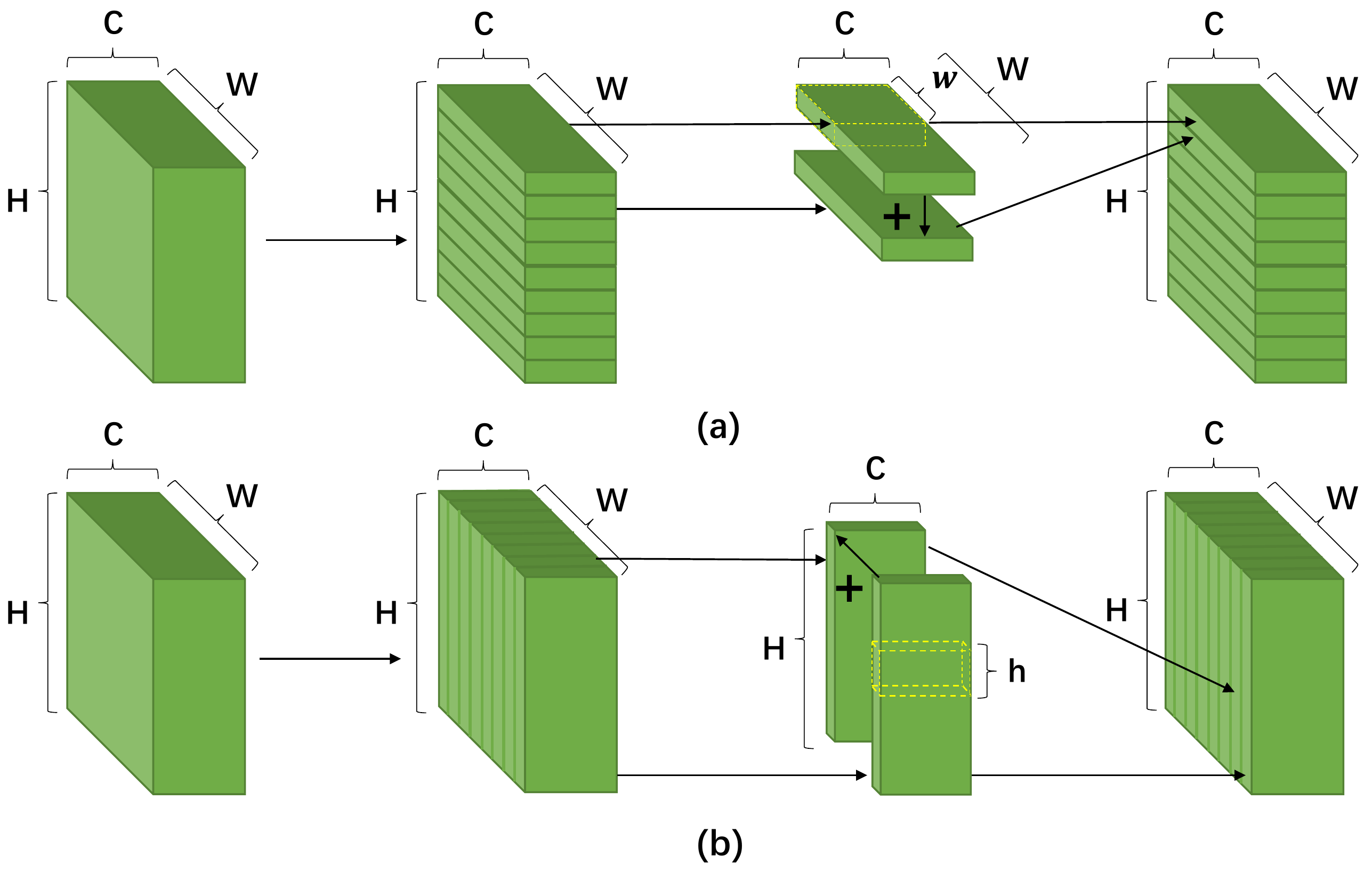}}
	\end{minipage}
	\caption{(a) Vertical slice convolution procedure (top-down). (b) Horizontal slice convolution procedure (right-left). The part with yellow dotted line is the convolutional filter.}
	\label{fig:hv}
\end{figure}

\begin{figure}[t]
	\begin{minipage}[b]{1.0\linewidth}
		\centering
		\centerline{\includegraphics[width=8cm]{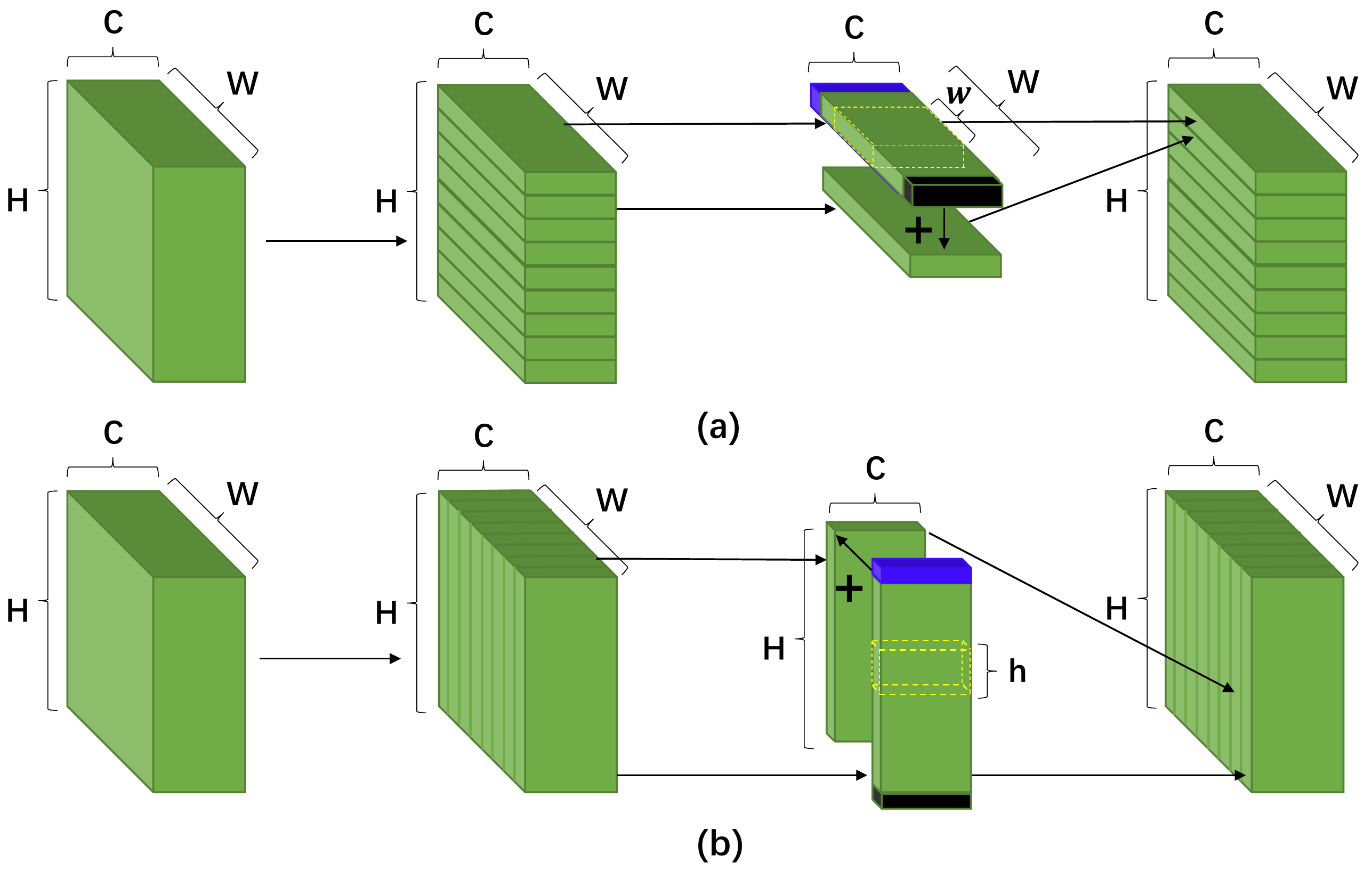}}
	\end{minipage}
	\caption{(a) Main diagonal slice convolution procedure (upper left-lower right). (b) Counter diagonal slice convolution procedure (upper right-lower left). The part with yellow dotted line is the convolutional filter. The black part of size $C\times1$ is abandoned. The blue part with the same size is the zero padding, which is used to fill the other side of the slice.}
	\label{fig:mbd}
\end{figure}

\begin{figure}[t]
	\begin{minipage}[b]{1.0\linewidth}
		\centering
		\centerline{\includegraphics[width=8cm]{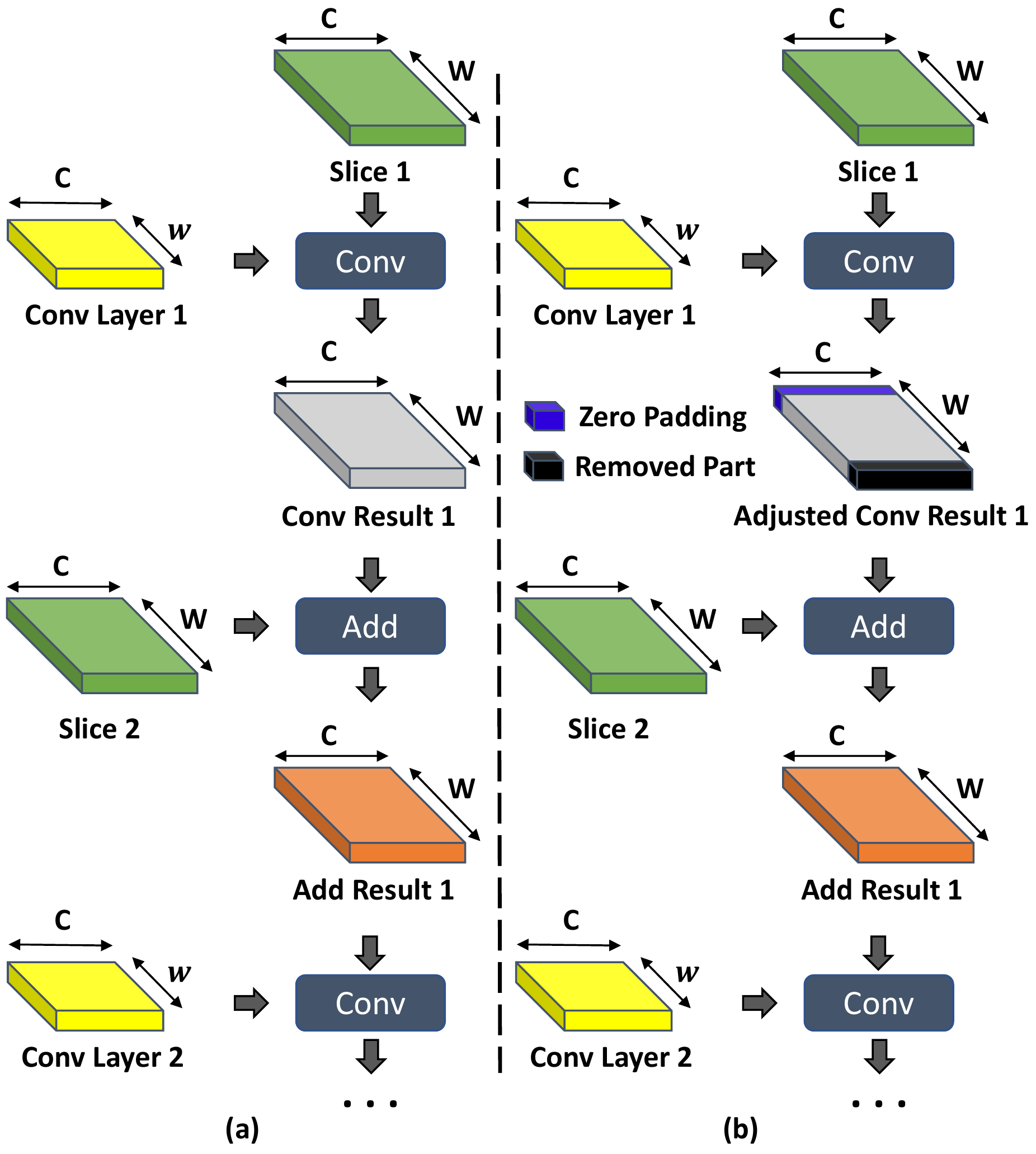}}
	\end{minipage}
	\caption{Detailed process of (top-down) horizontal (a) and (upper left-lower right) vertical (b) slice convolution.}
	\label{fig:detail}
\end{figure}

SCNN \cite{pan2018spatial} firstly adopts slice convolution to extract structural information for lane marker extraction task. The slice convolution used in SCNN is named as naive slice convolution in this paper. It could be divided into two types in terms of computation/sum direction, vertical slice convolution and horizontal slice convolution. The whole procedure of vertical and horizontal slice convolution is shown in Fig. \ref{fig:hv}.

Take the vertical slice convolution as an example. The input is the output of previous backbone network. It is a 3D tensor of size $C \times H \times W$, where $C$, $H$, and $W$ represent channel, row, and column number respectively. The tensor would be cut into $H$ slices, each of which becomes a slice of size $C \times 1 \times W$. Then a convolutional filter of size $C \times 1 \times w$, where $w$ is the width of filter, is applied to the first slice. Afterwards, the output of slice convolution is added to the second slice to become a new one. This is different with the traditional convolution, in which the output of a convolution layer is fed into the next layer. The new slice is sent to the next convolution layer and the process is repeated until the last slice tensor is handled. The detailed process is shown in Fig. \ref{fig:detail}(a). The first slice of input tensor and all adding results would be stacked to form the final output of vertical slice convolution.

There are two types of vertical slice convolution, top-down and bottom-up, depends on the location of the first slice and computation direction. The one we discussed above is the top-down type. If we start with the last slice of the input tensor, and compute upwardly, then we get the bottom-up slice convolution. Horizontal slice convolution is similar with this, except that the 3D tensor is cut into $W$ slices of size $C \times H \times 1$ and the convolution filter size becomes $C \times h \times 1$, where $h$ is the height of filter. It also contains two types, right-left and left-right.

\subsection{Multidirectional Slice Convolution}
\label{osc}

\begin{figure}[t]
	\begin{minipage}[b]{1.0\linewidth}
		\centering
		\centerline{\includegraphics[width=8cm]{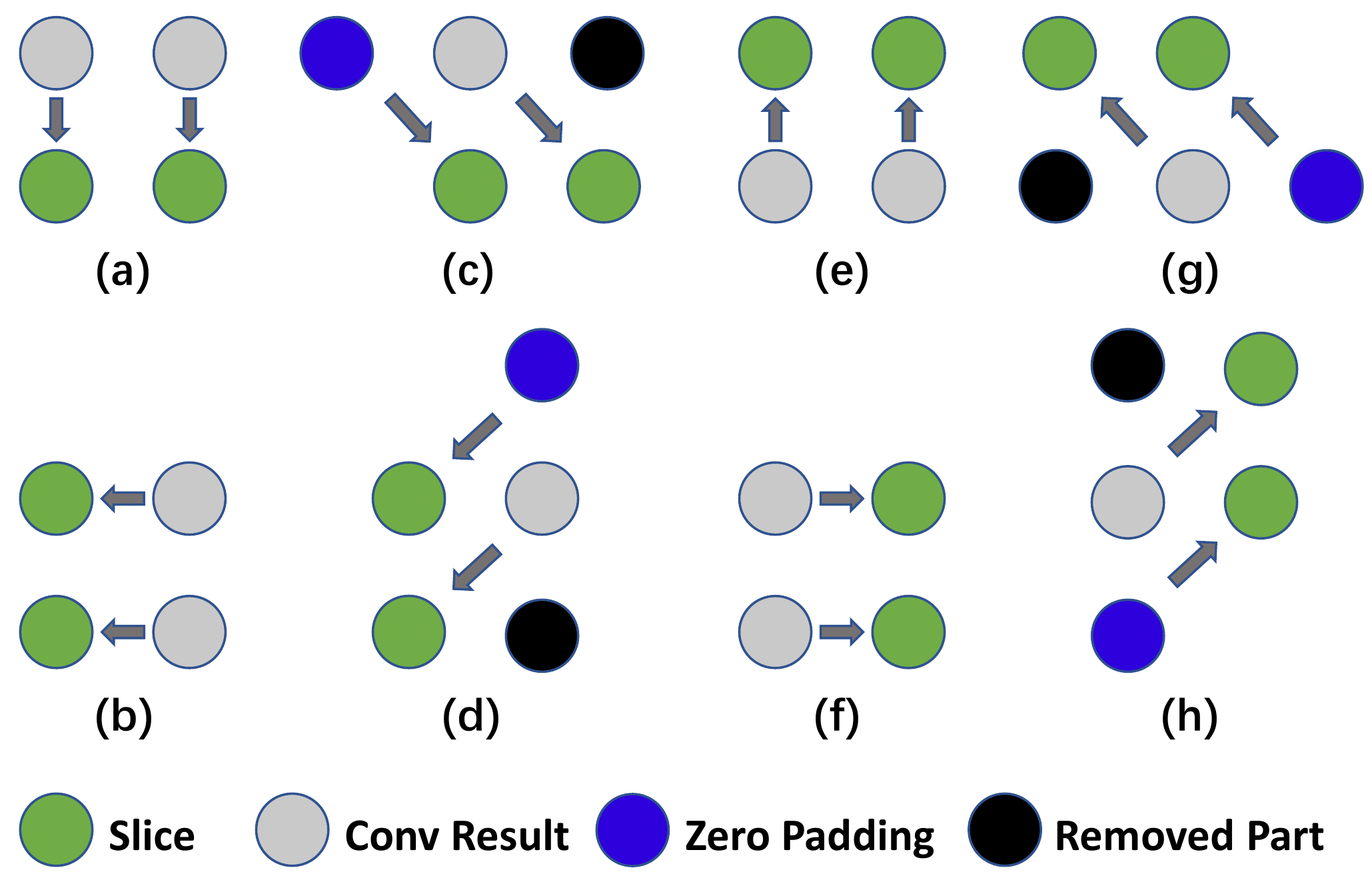}}
	\end{minipage}
	\caption{Computation directions of different slice convolution operations. (a) Vertical slice convolution (top-down). (b) Horizontal slice convolution (right-left). (c) Main diagonal slice convolution (upper left-lower right). (d) Counter diagonal slice convolution (upper right-lower left). (e) Vertical slice convolution (bottom-up). (f) Horizontal slice convolution (left-right). (g) Main diagonal slice convolution (lower right-upper left). (h) Counter diagonal slice convolution (lower left-upper right).}
	\label{fig:direct}
\end{figure}

Though SCNN uses slice convolution to extract structural information, it only captures the structural relation across rows and columns. This is not enough for lane marker extraction task, since most lanes present no exact horizontal or vertical shape in the front-view DVS images.

To tackle the problem, we propose Multidirectional Slice Convolution module. MSC module consists of slice convolution across all main orientations, including horizontal direction, vertical direction and diagonal direction. Hence, MSC is able to capture relation across these directions and extract more proper features for lanes. As a result, it can deal with occlusion situation much better than traditional convolutional layer. Following experimental results would validate its effectiveness in this aspect.

For diagonal slice convolution, the procedure is presented in Fig. \ref{fig:mbd}. This convolution contains operations along the main diagonal and counter diagonal directions. Take the case of main diagonal direction. Considering a 3D tensor which is split into $H$ slices of size $C \times 1 \times W$, a convolutional filter of size $C \times 1 \times w$, where $w$ is the width of filter, is applied to the first slice. The preprocess step is the same with vertical slice convolution. Then the output of slice convolution is shifted inside by one pixel along $W$ axis. The vector of size $C \times 1$ on the inner side, i.e., the black part, is abandoned. The outer side of the slice is filled with a zero vector/padding of same size, i.e., the blue part. The processed slice is added to the second slice next. The process continues until the last slice. The detailed process is shown in Fig. \ref{fig:detail}(b). The first slice of input tensor and all adding results would be stacked to form the final output of diagonal slice convolution.

There are also two types of main diagonal convolution, from upper left to lower right and the direction backwards, depending on the first slice location and computation direction. Counter diagonal slice convolution is alike. Its preprocess step is the same with horizontal slice convolution. The output of slice convolution is shifted downwards by one pixel along $H$ axis. The vector of size $C \times 1$ on the downward side, i.e., the black part, is abandoned. The upper side of the slice is filled with a zero vector of same size, i.e., the blue part. It includes two types, from upper right to lower left and the backward direction, too. Fig. \ref{fig:direct} shows computation directions of these slice convolution operations discussed so far.

The slice convolution process could be formulated as:
\begin{equation}
\label{E0}
X_i'=
\begin{cases}
X_i& {i=1}\\
X_i\oplus F(X_{i-1}'\otimes K)& {1< i\leq N},
\end{cases}
\end{equation}
where $X_i$ stands for the $i$-th slice of the input tensor, $X_i'$ denotes the $i$-th updated slice, $K$ represents the convolutional kernel, and $F$ is the ReLU activation function. $N$ equals to $H$ for vertical and main diagonal slice convolution, or $W$ for horizontal and counter diagonal slice convolution. $\otimes$ denotes convolution operation, and $\oplus$ denotes adding the $i$-th slice of the input tensor to the $(i-1)$th updated slice. This is a direct sum for horizontal and vertical slice convolution or a shifted sum for diagonal slice convolution.

\begin{figure}[t]
	\begin{minipage}[b]{1.0\linewidth}
		\centering
		\centerline{\includegraphics[width=8cm]{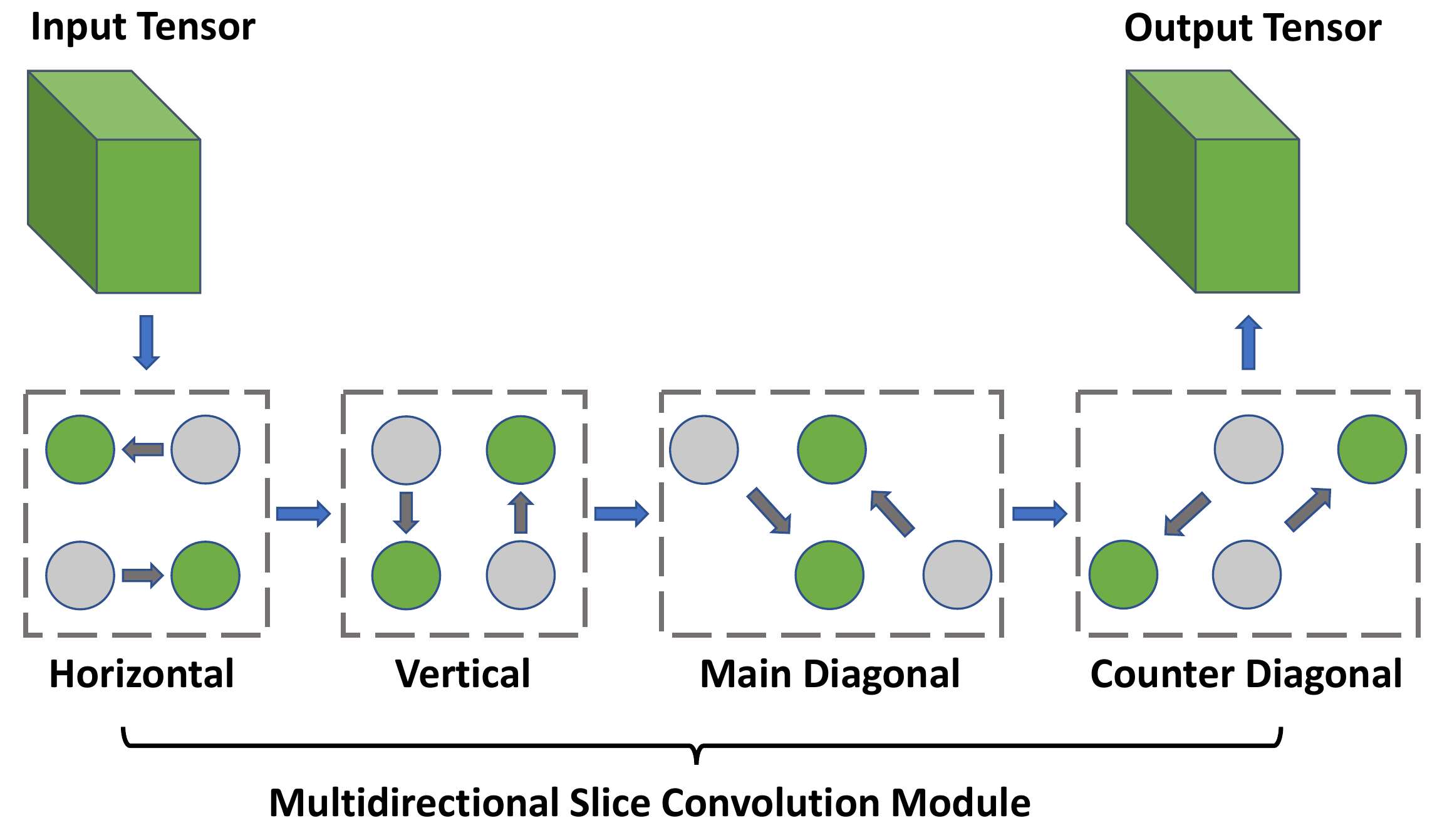}}
	\end{minipage}
	\caption{The Multidirectional Slice Convolution module structure. MSC module is composed of 4 types of slice convolution operations connected in series, and each of them contains 2 subtypes.}
	\label{fig:msc}
\end{figure}

Based on these slice convolution operations, we build the Multidirectional Slice Convolution module, as shown in Fig. \ref{fig:msc}. It is made up of 4 types of slice convolution operations, and each of them has 2 subtypes. These operations are connected in series to capture structural information along 8 directions. The input tensor and the output tensor has the same size.

\subsection{Structure-Aware Network}
With MSC module, Structure-Aware Network is proposed. The network is composed of three parts, the backbone network, MSC module, and the upsampling layer. Fig. \ref{fig:all} presents the whole structure of the network.

The backbone network is used to extract features from input image. Following SCNN, we adopt the DeepLab LargeFOV \cite{chen2017deeplab} variant for fair comparison. The MSC module is able to capture structural information from 8 directions, which is very helpful for the lane marker extraction task with DVS images. The output of MSC module is fed into a $1 \times 1$ convolution layer to make the channel number fit the class number. Then an upsampling layer is applied to restore the resolution of the output to be the same with the original input image, because the input image has been downsampled in the backbone network.

%% file: source/05exp.tex
\section{Experiments}

\subsection{Experimental Settings}
\label{sec:evaluate}
\textbf{Dataset Setting.} We conduct evaluations with two kinds of lane marker extraction methods, general semantic segmentation methods and specialized method for lane marker extraction task. The training set and validation set are used together to train these models to fully utilize the labeled data, and test set to evaluate their performance.

\textbf{Training Details.}
We adopt one Titan Xp GPU with 12 GB memory to train all models. The batch size is set as 4. The optimization algorithm is stochastic gradient descent (SGD) with  momentum. The value of momentum is set as 0.9. Poly learning rate policy is adopted to adjust learning rate, which reduces the learning rate per iteration. The process could be written as:
\begin{equation}
\label{Eq1}
LR=initial\;LR \times (1-\frac{current\;iter}{max\;iter})^{power},
\end{equation}
where $LR$ is the current learning rate, $initial\;LR$ is the initial learning rate, $current\;iter$ is the current iteration step, and $max\;iter$ is the max iteration step. The $initial\;LR$ is set as 0.01 and $power$ is set as 0.9. For all models, $max\;iter$ is set to 50,000 in our experiments to make sure these networks converge, and the corresponding number of training epoch is 74. The normalized cross entropy function is set as loss function. Considering the imbalanced pixels between background and lanes, the weight of background loss is set as 0.4. It could be formulated as:
\begin{equation}
\label{Eq2}
L=\lambda_bL_b+\lambda_lL_l,
\end{equation}
where $L_b$, $L_l$ are cross entropy loss functions of background and lanes, and $\lambda_b$, $\lambda_l$ equals 0.4, 1.0, respectively.

\begin{figure*}[t]
	\begin{minipage}[b]{1.0\linewidth}
		\centering
		\centerline{\includegraphics[width=16cm]{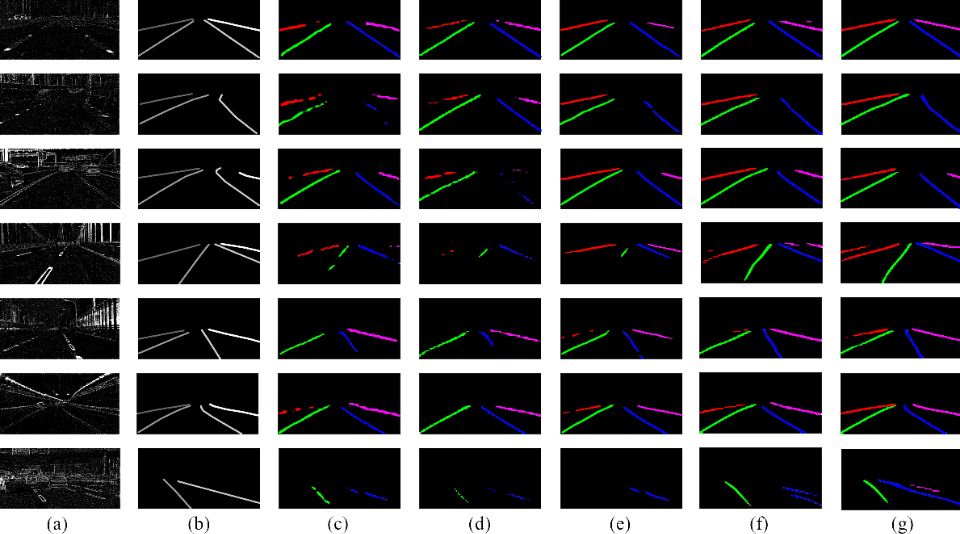}}
	\end{minipage}
	\caption{Visual comparison of lane marker extraction methods. (a) shows input images. (b) shows the corresponding label. (c-g) show results of FCN, DeepLabv3, RefineNet, SCNN and ours. The results are painted in color for better visual effect. Our method outperforms other methods significantly, especially on the connectivity of lanes. This demonstrates its ability to deal with occlusion and broken situations.}
	\label{fig:rlt2}
\end{figure*}

\textbf{Metrics.} Following the task definition of multi-class semantic segmentation, we choose two metrics to compare numerically. Specifically, we adopt widely used \emph{F1 score} (F1) and \emph{intersection over union} (IoU).

F1 is defined as
\begin{equation}
\label{E10}
{\rm F1}=2\times \frac{\rm Precision \times Recall}{\rm Precision+ Recall},
\end{equation}
and
\begin{equation}
{\rm Precision}=\frac{TP}{TP+FP},\;{\rm Recall}=\frac{TP}{TP+FN},
\end{equation}
where $TP,\;FP,\;TN,\;FN$ are the number of true positive, false positive, true negative, false negative separately. They all count the number of \emph{pixels}. IoU is defined as:
\begin{equation}
\label{Eq4}
{\rm IoU}(P_{m},P_{gt})= \frac{\mathbb{N}(P_{m}\cap P_{gt})}{\mathbb{N}(P_{m}\cup P_{gt})}, 
\end{equation} 
where $P_{m}$ is the prediction pixels set and $P_{gt}$ is the ground truth pixels set. $\cap$ and $\cup$ mean the intersection and union set respectively. $\mathbb{N}$ denotes the number of pixels in the intersection or union set. F1 and IoU are calculated across all five classes.

\subsection{Ablation Study}
In this section, we would make detailed ablation study to analyze our proposed approach comprehensively. 

\textbf{Multidirectional Slice Convolution.} We study the effect of directions in MSC at first. SANet with different directional slice convolution has been tested, and the results are shown in Tab. \ref{tab:osc}. The convolutional kernel size, i.e., the width or height of the kernel illustrated in Sec. \ref{osc}, is set to 9. Baseline model is the DeepLab LargeFOV \cite{chen2017deeplab} variant used in SCNN without any directional slice convolution. V, H, MD, CD denote slice convolution along the vertical, horizontal, main diagonal and counter diagonal direction separately. VH, MCD, MSC represent vertical and horizontal, main and counter diagonal, all of them respectively.

Tab. \ref{tab:osc} shows that the performance increases with more directions. Note that SANet\_MD with single direction performs better than SANet\_VH with both horizontal and vertical directions. This shows that oblique direction matters for lane marker extraction task. To verify that the improvement is brought by MSC, we add 8 extra 3x3 convolution layers to the baseline model to make it has similar number of parameters with our proposed model SANet\_MSC. The results demonstrate that SANet\_MSC outperforms ExtraConv\_8 significantly, which verifies the effectiveness of proposed MSC.

\begin{table}[h]
	\caption{Experimental results of multidirectional slice convolution.}
	\begin{center}
		\begin{tabular}{l|cc}
			\hline
			Models & Mean F1(\%)& Mean IoU(\%)\\
			\hline
			\hline
			Baseline & 72.43& 58.95\\
			SANet\_V & 73.82&60.43\\
			SANet\_H & 73.67&60.27 \\
			SANet\_MD & 73.91&60.52\\
			SANet\_CD & 73.82&60.43 \\
			SANet\_VH & 73.85&60.44\\
			SANet\_MCD & 73.96&60.59\\
			SANet\_MSC & \textbf{74.21}&\textbf{60.86} \\
			ExtraConv\_8 & 73.78&60.38\\
			\hline
		\end{tabular}
		\label{tab:osc}
	\end{center}
	\vspace{-0.5cm}
\end{table}

\textbf{Convolutional Kernel Size.} We then investigate the influence of slice convolutional kernel size, as presented in Tab. \ref{tab:kernel}. We use SANet with MSC module in this experiment. It shows that larger kernel size would be beneficial. Note that the model with kernel size of 11 performs worse than kernel size of 9. We argue this may be related to the relative size between feature map and kernel size.

\begin{table}[h]
	\caption{Experimental results of convolutional kernel size.}
	\begin{center}
		\begin{tabular}{l|ccccc}
			\hline
			Kernel Size & 3& 5& 7& 9& 11\\
			\hline
			\hline
			Mean F1(\%) & 73.76&73.85&74.18&\textbf{74.21}&74.10 \\
			Mean IoU(\%) & 60.36&60.47& 60.83&\textbf{60.86}&60.73\\
			\hline
		\end{tabular}
		\label{tab:kernel}
	\end{center}
	\vspace{-0.5cm}
\end{table}

\subsection{Evaluation on DET}
\textbf{Lane Marker Extraction Baselines.} We benchmark typical lane marker extraction methods, including general semantic segmentation based method, like FCN \cite{fcn}, DeepLabv3 \cite{deeplabv3}, RefineNet \cite{refinenet}, and specialized method for lane marker extraction task, SCNN \cite{pan2018spatial}. FCN, RefineNet and DeepLabv3 are widely used semantic segmentation methods for general computer vision problems. FCN is the first work taking semantic segmentation as pixel-level classification task. It contains a fully convolutional neural network and takes use of skip connections to combine shallow layer output with deep one. To utilize image-level global context, DeepLabv3 introduces atrous spatial pyramid pooling with global pooling operation. To make high-resolution prediction, RefineNet exploits information along the down-sampling process explicitly with long-range residual connections. 

SCNN is a specialized model for lane marker extraction task which has been introduced in Sec.\ref{sec:lane}.  SCNN achieves state-of-the-art performance on TuSimple \cite{tusimple} dataset. Tab.\ref{tab:exp} shows the results of lane marker extraction baselines and our method. Mean F1 ($\%$) refers to the average F1 score and Mean IoU ($\%$) represents the IoU of all classes.The best values are in bold and the second best values are underlined. Fig.\ref{fig:rlt2} shows visual results of these methods. 

\begin{table}[h]
	\caption{Evaluation results of lane marker extraction methods on DET.}
	\begin{center}
		\begin{tabular}{l|cc}
			\hline
			Methods &Mean F1(\%) & Mean IoU(\%)\\
			\hline
			\hline
			FCN \cite{fcn} & 62.32 & 49.04 \\
			DeepLabv3 \cite{deeplabv3}& 61.26 & 48.54\\
			RefineNet \cite{refinenet}& 64.34&50.97 \\
			SCNN \cite{pan2018spatial}& \underline{73.85}&\underline{60.44} \\
			Ours & \textbf{74.21}&\textbf{60.86}\\
			\hline
		\end{tabular}
		\label{tab:exp}
	\end{center}
	\vspace{-0.5cm}
\end{table}

Tab.\ref{tab:exp} shows that our SANet and SCNN outperforms other semantic segmentation methods. This is reasonable since FCN, DeepLabv3 and RefineNet are general semantic segmentation methods, and they do not contain special modules designed for lane marker extraction task. Structural feature or prior information is not applied either, which is critical for lane marker extraction task. SCNN adopts slice-by-slice convolution to capture continuous structure. This special module is really helpful for this task. 

Our proposed SANet with MSC module has achieved the best performance on DET dataset. It outperforms SCNN by 0.42\% on the strict metric, \emph{Mean IoU}, which is an obvious improvement for lane marker extraction task based on semantic segmentation. This demonstrates the effectiveness of our method.

%% file: source/06conc.tex
\section{Conclusion}
In this paper, a high-resolution DVS dataset for lane marker extraction task is constructed at first. It consists of the raw event data, accumulated images and corresponding labels. 5,424 event-based images with resolution of 1280$\times$800 pixels are extracted from 5 hours of event streams with sampling rate of MHz. To provide a comprehensive labeled pairs, two types of annotations for lanes in the images are given, multi-class format and binary format. Then the structure-aware network for lane marker extraction with DVS images is proposed, which can capture directional information extensively. We evaluate our proposed network with other state-of-the-art lane marker extraction models and analyze the results on DET dataset. Experimental results demonstrate that our method outperforms these methods on challenging datasets, which also verifies its high adaptability. Although our study focuses on the DVS images constructed from event stream, which makes it feasible to take use of existing methods designed for images, we would continue to explore techniques that directly take event stream as input to fully utilize the characteristic of event in the future.

%% file: my.bbl
\begin{thebibliography}{10}
\providecommand{\url}[1]{#1}
\csname url@samestyle\endcsname
\providecommand{\newblock}{\relax}
\providecommand{\bibinfo}[2]{#2}
\providecommand{\BIBentrySTDinterwordspacing}{\spaceskip=0pt\relax}
\providecommand{\BIBentryALTinterwordstretchfactor}{4}
\providecommand{\BIBentryALTinterwordspacing}{\spaceskip=\fontdimen2\font plus
\BIBentryALTinterwordstretchfactor\fontdimen3\font minus
  \fontdimen4\font\relax}
\providecommand{\BIBforeignlanguage}[2]{{%
\expandafter\ifx\csname l@#1\endcsname\relax
\typeout{** WARNING: IEEEtran.bst: No hyphenation pattern has been}%
\typeout{** loaded for the language `#1'. Using the pattern for}%
\typeout{** the default language instead.}%
\else
\language=\csname l@#1\endcsname
\fi
#2}}
\providecommand{\BIBdecl}{\relax}
\BIBdecl

\bibitem{zhu2016traffic}
Z.~Zhu, D.~Liang, S.~Zhang, X.~Huang, B.~Li, and S.~Hu, ``Traffic-sign
  detection and classification in the wild,'' in \emph{Proc. IEEE Conf. Comput.
  Vis. Pattern Recog.}, 2016, pp. 2110--2118.

\bibitem{li2018scale}
J.~Li, X.~Liang, S.~Shen, T.~Xu, J.~Feng, and S.~Yan, ``Scale-aware fast r-cnn
  for pedestrian detection,'' \emph{{IEEE} Trans. Multimedia}, vol.~20, no.~4,
  pp. 985--996, 2018.

\bibitem{tian2015pedestrian}
Y.~Tian, P.~Luo, X.~Wang, and X.~Tang, ``Pedestrian detection aided by deep
  learning semantic tasks,'' in \emph{Proc. IEEE Conf. Comput. Vis. Pattern
  Recog}, 2015, pp. 5079--5087.

\bibitem{borkar2012novel}
A.~Borkar, M.~Hayes, and M.~T. Smith, ``A novel lane detection system with
  efficient ground truth generation,'' \emph{{IEEE} Trans. Intell. Transp.
  Syst.}, vol.~13, no.~1, pp. 365--374, 2012.

\bibitem{deusch2012random}
H.~Deusch, J.~Wiest, S.~Reuter, M.~Szczot, M.~Konrad, and K.~Dietmayer, ``A
  random finite set approach to multiple lane detection,'' in \emph{Proc. IEEE
  Intell. Transp. Syst.}, 2012, pp. 270--275.

\bibitem{hur2013multi}
J.~Hur, S.-N. Kang, and S.-W. Seo, ``Multi-lane detection in urban driving
  environments using conditional random fields,'' in \emph{Proc. IEEE Intell.
  Vehicles Symp.}, 2013, pp. 1297--1302.

\bibitem{jung2013efficient}
H.~Jung, J.~Min, and J.~Kim, ``An efficient lane detection algorithm for lane
  departure detection,'' in \emph{Proc. IEEE Intell. Vehicles Symp.}, 2013, pp.
  976--981.

\bibitem{tan2014novel}
H.~Tan, Y.~Zhou, Y.~Zhu, D.~Yao, and K.~Li, ``A novel curve lane detection
  based on improved river flow and ransa,'' in \emph{Proc. IEEE Intell. Transp.
  Syst}, 2014, pp. 133--138.

\bibitem{wu2014lane}
P.-C. Wu, C.-Y. Chang, and C.~H. Lin, ``Lane-mark extraction for automobiles
  under complex conditions,'' \emph{Pattern Recognit.}, vol.~47, no.~8, pp.
  2756--2767, 2014.

\bibitem{gopalan2012learning}
R.~Gopalan, T.~Hong, M.~Shneier, and R.~Chellappa, ``A learning approach
  towards detection and tracking of lane markings,'' \emph{{IEEE} Trans.
  Intell. Transp. Syst.}, vol.~13, no.~3, pp. 1088--1098, 2012.

\bibitem{kim2014robust}
J.~Kim and M.~Lee, ``Robust lane detection based on convolutional neural
  network and random sample consensus,'' in \emph{Proc. Int. Conf. Neural Inf.
  Process.}, 2014, pp. 454--461.

\bibitem{huval2015empirical}
B.~Huval, T.~Wang, S.~Tandon, J.~Kiske, W.~Song, J.~Pazhayampallil,
  M.~Andriluka, P.~Rajpurkar, T.~Migimatsu, R.~Cheng-Yue \emph{et~al.}, ``An
  empirical evaluation of deep learning on highway driving,''
  \emph{arXiv:1504.01716}, 2015.

\bibitem{he2016accurate}
B.~He, R.~Ai, Y.~Yan, and X.~Lang, ``Accurate and robust lane detection based
  on dual-view convolutional neutral network,'' in \emph{Proc. IEEE Intell.
  Vehicles Symp.}, 2016, pp. 1041--1046.

\bibitem{li2017deep}
J.~Li, X.~Mei, D.~Prokhorov, and D.~Tao, ``Deep neural network for structural
  prediction and lane detection in traffic scene,'' \emph{{IEEE} Trans. Neural
  Netw. Learn. Syst.}, vol.~28, no.~3, pp. 690--703, 2017.

\bibitem{lee2017vpgnet}
S.~Lee, J.~Kim, J.~Shin~Yoon, S.~Shin, O.~Bailo, N.~Kim, T.-H. Lee,
  H.~Seok~Hong, S.-H. Han, and I.~So~Kweon, ``Vpgnet: Vanishing point guided
  network for lane and road marking detection and recognition,'' in \emph{Proc.
  IEEE Int. Conf. Comput. Vis.}, 2017, pp. 1947--1955.

\bibitem{binas2017ddd17}
J.~Binas, D.~Neil, S.-C. Liu, and T.~Delbruck, ``{DDD17}: End-to-end davis
  driving dataset,'' \emph{arXiv:1711.01458}, 2017.

\bibitem{valeiras2018event}
D.~R. Valeiras, X.~Clady, S.-H. Ieng, and R.~Benosman, ``Event-based line
  fitting and segment detection using a neuromorphic visual sensor,''
  \emph{{IEEE} Trans. Neural Netw. Learn. Syst.}, vol.~30, no.~4, pp. 1--13,
  2018.

\bibitem{cohen2018spatial}
G.~Cohen, S.~Afshar, G.~Orchard, J.~Tapson, R.~Benosman, and A.~van Schaik,
  ``Spatial and temporal downsampling in event-based visual classification,''
  \emph{{IEEE} Trans. Neural Netw. Learn. Syst.}, vol.~29, no.~10, pp. 1--15,
  2018.

\bibitem{camunas2017event}
L.~A. Camu{\~n}as-Mesa, T.~Serrano-Gotarredona, S.-H. Ieng, R.~Benosman, and
  B.~Linares-Barranco, ``Event-driven stereo visual tracking algorithm to solve
  object occlusion,'' \emph{{IEEE} Trans. Neural Netw. Learn. Syst.}, vol.~29,
  no.~9, pp. 4223--4237, 2017.

\bibitem{mueggler2017event}
E.~Mueggler, H.~Rebecq, G.~Gallego, T.~Delbruck, and D.~Scaramuzza, ``The
  event-camera dataset and simulator: Event-based data for pose estimation,
  visual odometry, and slam,'' \emph{Int. J. Robot. Res.}, vol.~36, no.~2, pp.
  142--149, 2017.

\bibitem{det}
W.~Cheng, H.~Luo, W.~Yang, L.~Yu, S.~Chen, and W.~Li, ``{DET}: A
  high-resolution dvs dataset for lane extraction,'' in \emph{Proc. IEEE Conf.
  Comput. Vis. Pattern Recog. Workshops}, 2019.

\bibitem{sironi2018hats}
A.~Sironi, M.~Brambilla, N.~Bourdis, X.~Lagorce, and R.~Benosman, ``{HATS}:
  Histograms of averaged time surfaces for robust event-based object
  classification,'' in \emph{Proc. IEEE Conf. Comput. Vis. Pattern Recog.},
  2018, pp. 1731--1740.

\bibitem{lagorce2017hots}
X.~Lagorce, G.~Orchard, F.~Galluppi, B.~E. Shi, and R.~B. Benosman, ``Hots: a
  hierarchy of event-based time-surfaces for pattern recognition,''
  \emph{{IEEE} Trans. Pattern Anal. Mach. Intell.}, vol.~39, no.~7, pp.
  1346--1359, 2017.

\bibitem{zhou2014object}
B.~Zhou, A.~Khosla, A.~Lapedriza, A.~Oliva, and A.~Torralba, ``Object detectors
  emerge in deep scene cnns,'' \emph{arXiv:1412.6856}, 2014.

\bibitem{szegedy2015going}
C.~Szegedy, W.~Liu, Y.~Jia, P.~Sermanet, S.~Reed, D.~Anguelov, D.~Erhan,
  V.~Vanhoucke, and A.~Rabinovich, ``Going deeper with convolutions,'' in
  \emph{Proc. IEEE Conf. Comput. Vis. Pattern Recog.}, 2015, pp. 1--9.

\bibitem{liu2015parsenet}
W.~Liu, A.~Rabinovich, and A.~C. Berg, ``Parsenet: Looking wider to see
  better,'' \emph{arXiv:1506.04579}, 2015.

\bibitem{visin2015renet}
F.~Visin, K.~Kastner, K.~Cho, M.~Matteucci, A.~Courville, and Y.~Bengio,
  ``Renet: A recurrent neural network based alternative to convolutional
  networks,'' \emph{arXiv:1505.00393}, 2015.

\bibitem{bell2016inside}
S.~Bell, C.~Lawrence~Zitnick, K.~Bala, and R.~Girshick, ``Inside-outside net:
  Detecting objects in context with skip pooling and recurrent neural
  networks,'' in \emph{Proc. IEEE Conf. Comput. Vis. Pattern Recog.}, 2016, pp.
  2874--2883.

\bibitem{pan2018spatial}
X.~Pan, J.~Shi, P.~Luo, X.~Wang, and X.~Tang, ``Spatial as deep: Spatial cnn
  for traffic scene understanding,'' in \emph{Proc. AAAI Conf. Artif. Intell.},
  2018.

\bibitem{li2017cifar10}
H.~Li, H.~Liu, X.~Ji, G.~Li, and L.~Shi, ``{CIFAR10-DVS}: an event-stream
  dataset for object classification,'' \emph{Front. Neurosci.}, vol.~11, p.
  309, 2017.

\bibitem{hu2016dvs}
Y.~Hu, H.~Liu, M.~Pfeiffer, and T.~Delbruck, ``{DVS} benchmark datasets for
  object tracking, action recognition, and object recognition,'' \emph{Front.
  Neurosci.}, vol.~10, p. 405, 2016.

\bibitem{vot2015}
``{VOT2015} benchmark,'' \url{http://www.votchallenge.net/vot2015/}.

\bibitem{tracking}
``Tracking dataset,'' \url{http://cmp.felk.cvut.cz/~vojirtom/dataset/tv77/}.

\bibitem{reddy2013recognizing}
K.~K. Reddy and M.~Shah, ``Recognizing 50 human action categories of web
  videos,'' \emph{Mach. Vis. Appl.}, vol.~24, no.~5, pp. 971--981, 2013.

\bibitem{griffin2007caltech}
G.~Griffin, A.~Holub, and P.~Perona, ``Caltech-256 object category dataset,''
  \emph{Technical report}, 2007.

\bibitem{aly2008real}
M.~Aly, ``Real time detection of lane markers in urban streets,'' in
  \emph{Proc. IEEE Intell. Vehicles Symp.}, 2008, pp. 7--12.

\bibitem{tusimple}
``{TuSimple Lane Detection Challenge},''
  \url{https://github.com/TuSimple/tusimple-benchmark/tree/master/doc/lane_detection}.

\bibitem{chiu2005lane}
K.-Y. Chiu and S.-F. Lin, ``Lane detection using color-based segmentation,'' in
  \emph{Proc. IEEE Intell. Vehicles Symp.}, 2005, pp. 706--711.

\bibitem{loose2009kalman}
H.~Loose, U.~Franke, and C.~Stiller, ``Kalman particle filter for lane
  recognition on rural roads,'' in \emph{Proc. IEEE Intell. Vehicles Symp.},
  2009, pp. 60--65.

\bibitem{teng2010real}
Z.~Teng, J.-H. Kim, and D.-J. Kang, ``Real-time lane detection by using
  multiple cues,'' in \emph{Proc. Int. Conf. Control Autom. Syst.}, 2010, pp.
  2334--2337.

\bibitem{lopez2010robust}
A.~L{\'o}pez, J.~Serrat, C.~Canero, F.~Lumbreras, and T.~Graf, ``Robust lane
  markings detection and road geometry computation,'' \emph{Int. J. Automot.
  Technol.}, vol.~11, no.~3, pp. 395--407, 2010.

\bibitem{fischler1981random}
M.~A. Fischler and R.~C. Bolles, ``Random sample consensus: a paradigm for
  model fitting with applications to image analysis and automated
  cartography,'' \emph{Commun. ACM}, vol.~24, no.~6, pp. 381--395, 1981.

\bibitem{zhou2010novel}
S.~Zhou, Y.~Jiang, J.~Xi, J.~Gong, G.~Xiong, and H.~Chen, ``A novel lane
  detection based on geometrical model and gabor filter,'' in \emph{Proc. IEEE
  Intell. Vehicles Symp.}, 2010, pp. 59--64.

\bibitem{lane_net}
D.~Neven, B.~De~Brabandere, S.~Georgoulis, M.~Proesmans, and L.~Van~Gool,
  ``Towards end-to-end lane detection: an instance segmentation approach,'' in
  \emph{Proc. IEEE Intell. Vehicles Symp.}, 2018, pp. 286--291.

\bibitem{kim2017end}
J.~Kim and C.~Park, ``End-to-end ego lane estimation based on sequential
  transfer learning for self-driving cars,'' in \emph{Proc. IEEE Conf. Comput.
  Vis. Pattern Recog. Workshops}, 2017, pp. 30--38.

\bibitem{chen2017deeplab}
L.-C. Chen, G.~Papandreou, I.~Kokkinos, K.~Murphy, and A.~L. Yuille, ``Deeplab:
  Semantic image segmentation with deep convolutional nets, atrous convolution,
  and fully connected crfs,'' \emph{{IEEE} Trans. Pattern Anal. Mach. Intell.},
  vol.~40, no.~4, pp. 834--848, 2017.

\bibitem{fcn}
J.~Long, E.~Shelhamer, and T.~Darrell, ``Fully convolutional networks for
  semantic segmentation,'' in \emph{Proc. IEEE Conf. Comput. Vis. Pattern
  Recog.}, 2015, pp. 3431--3440.

\bibitem{deeplabv3}
L.-C. Chen, G.~Papandreou, F.~Schroff, and H.~Adam, ``Rethinking atrous
  convolution for semantic image segmentation,'' \emph{arXiv:1706.05587}, 2017.

\bibitem{refinenet}
G.~Lin, A.~Milan, C.~Shen, and I.~Reid, ``Refinenet: Multi-path refinement
  networks for high-resolution semantic segmentation,'' in \emph{Proc. IEEE
  Conf. Comput. Vis. Pattern Recog.}, 2017, pp. 1925--1934.

\end{thebibliography}
